\documentclass{article} 

\usepackage{arxiv}

\usepackage[utf8]{inputenc}
\usepackage[T1]{fontenc}
\usepackage{hyperref}
\usepackage{url}
\usepackage{amsfonts} 
\usepackage{graphicx}
\usepackage{natbib}
\usepackage{doi}
\usepackage{graphicx}
\usepackage{float}
\usepackage{caption}
\usepackage{subcaption}
\usepackage{algorithmic}
\usepackage{pgfplots}
\usepackage{amsmath}
\usepackage{amssymb}
\usepackage{amsthm}
\usepackage[toc]{appendix}

\usepackage{booktabs}
\usepackage{longtable}
\usepackage{mathrsfs}
\usepackage{epsfig, epstopdf}
\usepackage[mathscr]{euscript}
\usepackage{lmodern}
\usepackage[scaled=0.85]{beramono}
\usepackage{multicol}
\usepackage{array}
\usepackage{color}
\usepackage{physics}
\usepackage{xcolor}
\usepackage{makecell}
\usepackage{mathtools}
\usepackage{bm}

\newtheorem{definition}{Definition}

\title{Phase-Aware Deep Learning with Complex-Valued CNNs for Audio Signal Applications}

\author{ Agrawal Naman \\
    Department of Statistics and Data Science \\
    National University of Singapore \\
	\texttt{naman.a@u.nus.edu} \\
}

\date{}

\renewcommand{\headeright}{}
\renewcommand{\undertitle}{}

\hypersetup{
pdftitle={Phase-Aware Deep Learning with Complex-Valued CNNs for Audio Signal Applications},
pdfauthor={Agrawal Naman}
}

\begin{document}
\maketitle

\begin{abstract}
This study explores the design and application of Complex-Valued Convolutional Neural Networks (CVCNNs) in the domain of audio signal processing, with a focus on preserving and utilizing phase information often neglected in real-valued networks. We begin by presenting the foundational theoretical concepts of CVCNNs, including complex convolutions, pooling layers, Wirtinger-based differentiation, and various complex-valued activation functions. These are complemented by critical adaptations of training techniques, including complex batch normalization and weight initialization schemes, to ensure stability in training dynamics. Empirical evaluations are conducted across three stages. First, CVCNNs are benchmarked on standard image datasets (MNIST, KMNIST, FMNIST), where they demonstrate competitive performance with real-valued CNNs, even under synthetic complex perturbations. Although our primary focus is audio signal processing, we first evaluate CVCNNs on image datasets to establish baseline performance and validate training stability before applying them to audio tasks. In the second experiment, we shift focus to audio classification using Mel-Frequency Cepstral Coefficients (MFCCs). CVCNNs trained on real-valued MFCCs slightly outperform real CNNs, while preserving phase in input workflows highlights challenges in exploiting phase without architectural modifications. Finally, a third experiment introduces Graph Neural Networks (GNNs) to model phase information via edge weighting explicitly, where the inclusion of phase yields measurable gains in both binary and multi-class genre classification, affirming its relevance when appropriately structured. These results collectively underscore the expressive capacity of complex-valued architectures and confirm phase as a meaningful and exploitable feature in audio processing applications. While current methods show promise, especially with activations like cardioid, future advances in phase-aware design will be essential to fully leverage the potential of complex representations in neural networks.
\end{abstract}


\section{Introduction}

The emergence of deep learning has significantly advanced the field of audio signal processing, enabling models to learn complex representations directly from raw data. Among these methods, CNNs have become particularly prominent due to their effectiveness in extracting hierarchical features from temporal and spectral representations of audio signals. CNN-based architectures have achieved state-of-the-art performance in a range of audio-related tasks, including speech recognition, acoustic scene classification, and sound source localization \cite{6857341, adavanne2017soundeventdetectionusing, hershey2017cnnarchitectureslargescaleaudio}.

The widespread adoption of deep learning frameworks such as TensorFlow, PyTorch, and Keras has further accelerated the development and deployment of CNNs. These tools allow for the modular design of neural network architectures incorporating multiple convolutional layers, activations, regularization techniques, and optimization strategies \cite{trabelsi2018deep}. As a result, CNNs have been successfully integrated into production systems for real-time and high-accuracy audio processing applications \cite{huang2018characterizing}.

Despite their success, conventional real-valued CNNs inherently discard or under-utilize phase information, a critical component in many signal processing tasks. This limitation motivates the exploration of CVCNNs, which extend real-valued networks by incorporating complex arithmetic into network operations. CVCNNs offer a natural framework to preserve and exploit both magnitude and phase information present in audio signals, potentially leading to more expressive and efficient representations.

Recent advances across various domains have demonstrated the significant benefits of using CVCNNs. In remote sensing, CVCNNs have shown exceptional performance in tasks that require preserving phase information, such as landform classification from InSAR data \cite{sunaga2019land}, forest height mapping \cite{wang2019forest}, and polarimetric SAR image classification \cite{zhang2017complex}. These applications leverage the ability of CVCNNs to capture both amplitude and phase components in geospatial signals, offering a more holistic representation than real-valued models. In medical imaging, CVCNNs have been successfully applied to reconstruct high-quality images from undersampled MRI data \cite{dedmari2018complex}, where phase information plays a critical role in de-aliasing. More recently, they have been used to enhance ultrasound beamforming with complex-valued recurrent networks \cite{zhang2025ultrasound}, demonstrating superior spatial resolution and noise robustness compared to traditional beamformers.

Within communication systems and signal processing, CVCNNs have proven effective in reducing bit error rates for MIMO wireless detection \cite{marseet2017application}, and in improving the quality and efficiency of radar imaging systems \cite{gao2019enhanced}. These models are well-suited for such tasks due to the complex-valued nature of wireless and radar signals. Beyond supervised learning, CVCNNs have also been adopted in unsupervised learning tasks. For instance, they have improved the accuracy and convergence speed of sleep stage classification from EEG signals without requiring labeled data \cite{zhang2018unsupervised}. In audio processing, phase-aware architectures such as the Deep Complex U-Net have outperformed traditional models in speech enhancement tasks \cite{choi2019phase}, thanks to their ability to jointly process magnitude and phase components of audio spectrograms.

Furthermore, emerging work in hardware acceleration and optical computing has pushed the envelope for real-time deployment. Hardware accelerators like TOPS-speed complex-valued convolutional chips \cite{bai2025tops} are being developed to support efficient complex-valued inference pipelines. These innovations open up possibilities for energy-efficient and high-speed CVCNN deployment in edge and embedded systems. Together, these works highlight the expressive power of CVCNNs in handling data with inherent phase structure, offering compelling evidence for their integration into a wider range of machine learning pipelines.

This paper investigates the potential of CVCNNs for audio signal processing, with an emphasis on tasks such as music genre classification. We first establish the mathematical foundations of CVCNNs, including complex convolutions, activation functions, batch normalization techniques, and weight initialization strategies, to support stable and expressive network training. Our study empirically evaluates CVCNNs in three key settings:
\begin{itemize}
    \item on standard image datasets (MNIST, KMNIST, FMNIST) to benchmark baseline performance and verify training stability under synthetic complex perturbations;
    \item on music genre classification using MFCCs, comparing real- and complex-valued inputs to assess the impact of incorporating phase information;
    \item in phase-aware configurations where phase is explicitly modeled GNNs 
\end{itemize}
Through these investigations, we demonstrate the conditions under which CVCNNs achieve competitive or superior performance relative to real-valued CNNs, highlight the benefits and limitations of phase-aware modeling, and provide evidence supporting the broader adoption of complex-valued architectures in machine learning applications.

\section{Theoretical Frameworks}

We start by examining the key components of the CVCNN architecture, such as complex-valued convolutions, activation functions, and parameter initialization. To contextualize their relevance, these characteristics are systematically compared with their counterparts in traditional real-valued CNNs, highlighting both structural similarities and functional differences.  We start this discussion with the core operation that underpins CNN architectures: convolution.

\subsection{Complex Valued Convolutions}

The convolution of two functions $f$ and $g$ is defined as the integral of the product of the two functions after one is reflected about the y-axis and shifted. Mathematically,
$$(f * g)(t) = f(t) * g(t) = \int_{-\infty}^\infty f(\tau) g(t - \tau) \; d \tau = \int_{-\infty}^\infty f(t - \tau) g(\tau) \; d \tau$$
For $f$ and $g$ defined over the set of integers $\mathbb{Z}$, the discrete convolution is defined as follows:
$$(f * g)[n] = f[n] * g[n] = \sum_{m = -\infty}^\infty f[m] g[n - m] = \sum_{m = -\infty}^\infty f[n - m] g[m]$$
The above notion can be easily extended to two or more variables, forming the basis for convolutions between matrices:
$$(f * g)[n_1, n_1] = f[n_1, n_2] * g[n_1, n_2] = \sum_{m_1 = -\infty}^\infty \; \sum_{m_2 = -\infty}^\infty f[m_1, m_2] g[n_1 - m_1, n_2 - m_2]$$
where $f$ is called the input and $g$ is called the kernel. This allows us to define convolutional operations between 2D matrices by taking $f(x, y) = A_1[x, y]$ and $g(x, y) = A_2[x, y]$. Now, given a complex valued input matrix $X = A_1 + iB_1$ and a complex valued kernel $W = A_2 + iB_2$, there complex valued convolution may be defined as follows \cite{trabelsi2018deep}:
$$W * X = (A_1 + iB_1) * (A_2 + iB_2) = (A_1 * A_2 - B_1 * B_2) + i(B_1 * A_2 + A_1 * B_2)$$
One way to understand the above extension would be to consider the matrix representation of complex numbers:
$$W * X = 
\left(
\begin{array}{cc}
    A_1 & -B_1 \\
    B_1 & A_1
\end{array}
\right) *
\left(
\begin{array}{cc}
    A_2 & -B_2 \\
    B_2 & A_2
\end{array}
\right)
=
\left(
\begin{array}{cc}
    A_1 * A_2 - B_1 * B_2 & -B_1 * A_2 - A_1 * B_2 \\
    B_1 * A_2 + A_1 * B_2 & A_1 * A_2 - B_1 * B_2
\end{array}
\right)
$$
$$=
\left(
\begin{array}{cc}
    \text{Re}(W * X) & -\text{Im}(W * X)  \\
    \text{Im}(W * X)  & \text{Re}(W * X) 
\end{array}
\right) =
(A_1 * A_2 - B_1 * B_2) + i(B_1 * A_2 + A_1 * B_2)
$$

This formulation allows us to extend the real-valued convolution operation into the complex domain in a mathematically consistent and computationally feasible manner. The matrix representation offers a valuable perspective by encoding complex multiplication as a structured operation on real-valued tensors, making it easier to implement complex-valued layers using existing real-valued operations in deep learning frameworks, while still maintaining the integrity of complex arithmetic. This structured yet flexible approach to defining complex convolutions forms the backbone of CVCNN architectures.

\subsection{Pooling Layers}

Pooling functions summarize the presence of features in local patches of feature maps produced after convolution operations. They serve as downsampling layers that reduce spatial dimensions and help suppress noise or irrelevant details in the input \cite{9849162}. The two most common strategies are max pooling and average pooling, which respectively compute the maximum or average value within a neighborhood of the feature map.

However, extending pooling operations to the complex domain poses unique challenges. In particular, the max operator is not naturally defined for complex numbers. As noted by Popa \cite{popa2017complex}, one workaround is to apply max pooling separately to the real and imaginary parts. Yet, this neglects the joint magnitude-phase structure of complex representations.

To address this, we adopt a more principled method where complex max pooling is performed based on the magnitude of the complex values, in line with \cite{guberman2016complex}. The corresponding phase is then recovered using the index selected by max pooling on the magnitudes. In contrast, complex average pooling is more straightforward and involves separately averaging the real and imaginary components. These formulations preserve the integrity of complex representations during spatial downsampling, enabling effective feature aggregation in CVCNNs.

\subsection{Complex Calculus Preliminaries}

Complex calculus extends the principles of real calculus into the complex plane and, therefore, constitutes an integral tool for our discussion on CVCNNs and their activations. Given an input $z = x + iy \in \mathbb{C}$, we can define a complex function as follows:
$$f: \mathbb{C} \rightarrow \mathbb{C}; \; x + iy \mapsto u(x, y) + iv(x, y)$$
where $u$ and $v$ are real valued functions in $x$ and $y$. 
\begin{definition}
    A complex function $f: \mathbb{C} \rightarrow \mathbb{C}$ is said to be real differentiable if $\pdv{u}{x}$ and $\pdv{u}{y}$ are well defined.
\end{definition}
\begin{definition}
    A complex function $f: \mathbb{C} \rightarrow \mathbb{C}$ is said to be complex-differentiable or analytic or holomorphic if it satisfies the Cauchy-Riemann equations:
    $$\left(\pdv{u}{x} = \pdv{v}{y}\right) \wedge \left(\pdv{u}{y} = -\pdv{v}{x}\right)$$
\end{definition}
Unlike in conventional real-valued neural networks, the use of complex-valued activation functions introduces additional mathematical challenges, particularly in the context of gradient-based optimization. In classical complex analysis, a function must be holomorphic (i.e., complex differentiable) to possess a valid derivative in the complex domain. However, many activation functions employed in CVCNNs do not satisfy the Cauchy-Riemann conditions and are therefore non-holomorphic (as we will see in Section \ref{seccva}). To address this issue, Wirtinger calculus (also known as $\mathbb{CR}$ calculus) is utilized. This framework allows differentiation of complex-valued functions that are not analytic by treating the real and imaginary components of complex variables as independent. Specifically, a function $f(z) = f(x + iy)$ is considered differentiable in the Wirtinger sense if it is differentiable for the real part $x$ and the imaginary part $y$ separately. This approach enables the application of standard optimization techniques, such as stochastic gradient descent, by computing gradients using partial derivatives with respect to $z$ and its conjugate $\bar{z}$, even when $f(z)$ is not holomorphic \cite{virtue2017better}. This, in turn, forms a crucial theoretical foundation for training complex-valued neural networks. It ensures that meaningful gradients can be computed for a broad class of activation functions and loss surfaces, enabling the effective backpropagation of errors during learning in CVCNNs. More detailed insights into Wirtinger calculus and how backpropagation can be implemented for complex-valued neural networks can be found in \cite{hammad2024comprehensivesurveycomplexvaluedneural, barrachina2023theoryimplementationcomplexvaluedneural, 9849162}.

\subsection{Complex Valued Activations} \label{seccva}

Activation functions play an important role in the functioning of neural networks by introducing non-linearity, enabling the model to capture complex relationships within the data. For CVCNNs, where both real and imaginary components are considered, the choice of activation function takes on added significance due to the inherently richer representational capacity afforded by complex numbers. Traditional real-valued activation functions, such as the rectified linear unit (ReLU) or sigmoid, find their complex-valued counterparts in functions that can effectively operate on both the magnitude and phase information present in complex numbers. The ideal complex activation function should strike a balance between enabling non-linearity, preserving phase information, and ensuring numerical stability during training. Moreover, it should be computationally efficient to facilitate any practical applications. Some of the popular complex-valued activation functions proposed in the literature are discussed in the subsequent subsections.

\subsubsection{\texorpdfstring{$\mathbb{C}$}{TEXT}ReLU}
The $\mathbb{C}$ReLU is the most natural extension of the ReLU activation to the complex domain, defined as follows:
$$\mathbb{C}\text{ReLU}: \mathbb{C} \rightarrow \mathbb{C}; \; z \mapsto \text{ReLU}(\text{Re}(z)) + i \text{ReLU}(\text{Im}(z))$$
When both real and imaginary components are simultaneously strictly positive or negative, the above function remains analytic by the Cauchy-Riemann equations \cite{trabelsi2018deep}.

\subsubsection{modReLU}

The modReLU activation is another variant of ReLU proposed by  (Arjovsky et al., 2015) \cite{arjovsky2016unitary}, defined as follows:
$$\text{modReLU}: \mathbb{C} \rightarrow \mathbb{C}; \; z \mapsto \text{ReLU}(|z| + b) \cdot \exp{(i \phi)} = \left\{
\begin{array}{cc}
    (|z| + b) \cdot \cfrac{z}{|z|} & |z| + b \geq 0  \\
    0 & |z| + b < 0
\end{array}
\right.$$
Where $b \in \mathbb{R}$ is a learnable parameter (bias). Unlike the $\mathbb{C}$ReLU, modReLU preserves the phase of the complex input even after the non-linear transformation (which otherwise could drastically affect the complex representation of the hidden states). However, this comes at the cost of a lack of differentiability as the modReLU activation no longer satisfies Cauchy Cauchy-Riemann equations. For instance, if $|z| + b \geq 0$:
$$x + iy \mapsto \left(x + \cfrac{bx}{\sqrt{x^2 + y^2}} \right) + i\left(y + \cfrac{by}{\sqrt{x^2 + y^2}} \right)$$
$$\pdv{u}{x} = \pdv{}{x}\left(x + \cfrac{bx}{\sqrt{x^2 + y^2}} \right) = 1 + \cfrac{by^2}{\sqrt{x^2 + y^2}}$$
$$ \pdv{v}{y} = \pdv{}{y}\left(y + \cfrac{by}{\sqrt{x^2 + y^2}} \right) = 
 1 + \cfrac{bx^2}{\sqrt{x^2 + y^2}} \neq \pdv{u}{x}$$

\subsubsection{zReLU}

Guberman (2016) \cite{guberman2016complex} proposed yet another variant of the complex ReLU function (which has been referred to as zReLU by some authors \cite{trabelsi2018deep}). zReLU is similar to $\mathbb{C}$ReLU, also offering a natural extension of the standard ReLU function as follows:
$$\text{zReLU}: \mathbb{C} \rightarrow \mathbb{C}; \; z \mapsto \left\{
\begin{array}{cc}
   z & (\text{Re}(z) \geq 0) \wedge (\text{Im}(z) \geq 0)  \\
    0 & (\text{Re}(z) < 0) \vee (\text{Im}(z) < 0)
\end{array}
\right.
\iff
\left\{
\begin{array}{cc}
   z & \phi_z \in \left[ 0, \frac{\pi}{2} \right]  \\
    0 & \phi_z \not\in \left[ 0, \frac{\pi}{2} \right]
\end{array}
\right.$$
Unlike the $\mathbb{C}$ReLU function, which separately applies the standard ReLU on both real and imaginary components, zReLU returns the original input only if both the real and imaginary components are positive. This function is also analytic for most of $\mathbb{C}$ except for the following set of points \cite{trabelsi2018deep}:
$$\left\{ z \in \mathbb{C} |  (\text{Re}(z) > 0) \wedge (\text{Im}(z) = 0)\right\} \cup \left\{ z \in \mathbb{C} |  (\text{Re}(z) = 0) \wedge (\text{Im}(z) > 0)\right\}$$

Due to these sharp discontinuities along the boundaries of the first quadrant, training neural networks with zReLU can become unstable and difficult, especially when gradients are required to flow smoothly through complex-valued layers. We propose a smoothed alternative to zReLU to mitigate this issue, called the smooth zReLU activation. It introduces a soft approximation using sigmoid functions to create a differentiable mask over the real and imaginary parts of the input: $z \cdot \sigma(\alpha \cdot \text{Re}(z)) \cdot \sigma(\alpha \cdot \text{Im}(z))$ where $\sigma(x) = \frac{1}{1 + e^{-x}}$ is the sigmoid function, and $\alpha > 0$ is a tunable sharpness parameter controlling the steepness of the transition. Typically, $\alpha$ is set to values like $1.0$ or $0.5$. As $\alpha \to \infty$, the function approaches the hard thresholding behavior of the original zReLU, while smaller $\alpha$ leads to a smoother and more gradient-friendly activation function, enabling better convergence during training.

\subsubsection{Others}

Besides the above commonly used complex activations, numerous others have been proposed in the literature. As per Liouville's Theorem, if $f(z)$ is a holomorphic function for all finite values of $z$ and is bounded for all $z \in \mathbb{C}$, then $f$ is a constant function. Thus, coming up with a complex-valued activation required a researcher to choose between differentiability and boundedness. As a result, most initial activation functions were non-analytic to preserve boundedness \cite{scardapane2018complexvalued}. One example was the complex extension of the usual hyperbolic tangent function as follows \cite{10.1162/089976603321891846}:
$$\tanh (z) = \cfrac{\exp{(z)} - \exp{(-z)}}{\exp{(z)} + \exp{(-z)}}$$
which has periodic singular points (that can be avoided to a large extent by carefully scaling the initial weights \cite{scardapane2018complexvalued}). Besides, similar to the $\mathbb{C}$ReLU activation, many other activation functions worked by applying a common real-valued activation $g(\cdot)$ (like tanh or sigmoid) separately on the real and imaginary components as follows:
$$g(z) := g(\text{Re}(z)) + ig(\text{Im}(z))$$
These functions (called split activations \cite{benvenuto1992complex, leung1991complex, benvenuto1991nonlinear}) constitute an important class of complex activations in modern-day research. Finally, another important complex activation called the complex cardioid was proposed by Patrick Virtue et al. (2017) \cite{virtue2017better}. It is defined as follows:
$$g(z) = \cfrac{z}{2} \left( 1 + \cos \phi_z\right)$$
Like modReLU, this method preserves phase information while also using it to modify the magnitude.

Thus, it is evident that researchers have explored various complex activation functions, ranging from algebraic formulations to transcendental functions inspired by mathematical operations on complex numbers. Each of these functions brings its own set of advantages and considerations, with some demonstrating superior performance in specific applications \cite{virtue2017better}. While this paper focuses on a selected subset of widely used and theoretically significant activation functions, a much broader taxonomy—including several recently proposed variants—is comprehensively discussed in \cite{hammad2024comprehensivesurveycomplexvaluedneural}, reflecting the growing body of research dedicated to designing and categorizing complex-valued activations.  As the field of complex-valued neural networks continues to advance, the development and evaluation of novel activation functions tailored to this domain represent an active area of research with a lot of potential in the modeling and processing of complex data.

\subsection{Complex Batch Normalization}

In traditional CNNs, the optimization and stability of training processes have been long-standing challenges. Common approaches often grapple with issues related to vanishing or exploding gradients, which hinder the convergence of the network. To address these concerns, Batch Normalization was introduced by Sergey Ioffe and Christian Szegedy in 2015 \cite{ioffe2015batch}. This method operates by normalizing the activations of each layer, specifically the mean and variance, over the mini-batch of data. By centering and scaling the activations, Batch Normalization reduces internal covariate shifts, leading to more stable gradients and faster convergence during training. This not only expedites the training process but also enables the use of higher learning rates, accelerating the optimization process.

Trabelsi et al. \cite{trabelsi2018deep} proposed a variant of batch normalization that extends to the complex domain. They have explained that the standardization of an array of complex numbers demands more than a simple translation and scaling process, where the mean is set to zero and the variance to one. This conventional normalization approach, while effective in univariate contexts, does not ensure equal variance in both the real and imaginary components. The resulting distribution may become elliptical by exhibiting substantial elongation or compression along different axes. Thus, we need a specialized approach to standardize complex-valued distributions to achieve circularity and provide a more accurate representation of the underlying data structures. For a complex-valued vector $x$, this involves the following computation:
$$\Tilde{x} = (\mathbf{V})^{-1/2} (x - \mathbb{E}(x))$$
$$\mathbf{V} = \left(
\begin{array}{cc}
    \mathbf{V}_{rr} &  \mathbf{V}_{ri} \\
    \mathbf{V}_{ir} & \mathbf{V}_{ii} 
\end{array}
\right) + \lambda \mathbf{I}=
\left(
\begin{array}{cc}
    \text{Cov}(\text{Re}(x), \text{Re}(x)) &  \text{Cov}(\text{Re}(x), \text{Im}(x)) \\
    \text{Cov}(\text{Im}(x), \text{Re}(x)) & \text{Cov}(\text{Im}(x), \text{Im}(x))
\end{array}
\right) + \lambda \mathbf{I}$$
$$\text{BN}(\Tilde{x}) = \gamma \Tilde{x} + \beta$$
where $\lambda \mathbf{I}$ has been added as Tikhonov regularization for ensuring the positive definiteness of $\mathbf{V}$, and $\gamma, \beta$ are learnable parameters for scaling after applying the standardization. This formulation ensures that the complex-valued input $x$ is standardized in a way that preserves the joint distribution of its real and imaginary parts. The mean $\mathbb{E}(x)$ is subtracted to center the data, while the covariance matrix $\mathbf{V}$ captures both intra-component (real-real, imaginary-imaginary) and inter-component (real-imaginary, imaginary-real) dependencies. Taking the inverse square root of $\mathbf{V}$ performs a whitening transformation that removes correlations and scales the data to have equal variance in all directions, achieving a circular (isotropic) distribution in the complex plane.

\subsection{Complex Weight Initialization}

In standard neural networks with real-valued weights, appropriate weight initialization is crucial for ensuring effective learning during training. Several techniques have been developed to set the initial values of the weights in a manner that facilitates convergence and improves performance. Traditional neural networks typically employ Normalized Xavier Weight Initialization or He Initialization, both of which aim to maintain a stable flow of gradients during backpropagation. However, when extending to complex-valued neural networks, the importance of careful initialization becomes even more pronounced. Unlike their real-valued counterparts, complex-valued networks are significantly more sensitive to the choice of initial weights, often resulting in unstable training dynamics or poor convergence when initialized improperly \cite{monning2018evaluation}. This heightened sensitivity necessitates the development of tailored initialization strategies that preserve the statistical properties of complex weights and ensure robust learning. In the following sections, we formally define a complex random variable and propose suitable complex-valued extensions of the Xavier and He initialization schemes.

\begin{definition}
    A complex random variables Z on a probability space $(\Omega, \mathcal{F}, P)$ is a function $Z: \Omega \rightarrow \mathbb{C}$ such that both its imaginary $(\Im (Z))$ and real $(\Re (Z))$) components are random variables on $(\Omega, \mathcal{F}, P)$.  Thus, complex random variables are equivalent to a bivariate real random vector $(X, Y)$, and its distribution is defined by the joint distribution of $(X, Y)$.
\end{definition}

\subsubsection{Normalized Xavier Weight Initialization}

For conventional neural networks, Xavier weight initialization involves sampling the weights for a layer from a uniform distribution \cite{pmlr-v9-glorot10a}. The parameters of the uniform distribution depend on the number of input nodes ($n$) and the number of outputs ($m$). In particular, the weights are sampled from the distribution:

$$Z \sim U \left[ - \sqrt{\cfrac{6}{n + m}},   \sqrt{\cfrac{6}{n + m}} \right]; \; f_Z (z) = 
\left\{
\begin{array}{cc}
     \cfrac{1}{2}\sqrt{\cfrac{n + m}{6}}, & \;\; \text{if }  - \displaystyle\sqrt{\cfrac{6}{n + m}} \leq z \leq \displaystyle\sqrt{\cfrac{6}{n + m}}  \\[13pt]
     0, & \;\; \text{else}
\end{array}
\right.
$$
The idea of the initialization algorithm was to maintain near-identical variances of its weight gradients across layers. The above algorithm can be extended for the complex uniform random variable: as we propose, one such extension samples complex weights from a circular uniform distribution, ensuring isotropic variance and rotational symmetry in the complex plane.

$$Z \sim CU\left[ - \sqrt{\cfrac{6}{n + m}},   \sqrt{\cfrac{6}{n + m}} \right]; \; f_Z (z) = 
\left\{
\begin{array}{cc}
     \cfrac{1}{\pi} \cdot \cfrac{n + m}{6}, & \;\; \text{if }  \left\| z \right\| \leq \displaystyle\sqrt{\cfrac{6}{n + m}}  \\[13pt]
     0, & \;\; \text{else}
\end{array}
\right.$$
This complex extension ensures that the weights are uniformly distributed within a disk in the complex plane, preserving isotropic variance and maintaining stable signal propagation in complex-valued neural networks. The process of sampling remains the same as that from bivariate distributions. An alternative approach, proposed in \cite{barrachina2023theoryimplementationcomplexvaluedneural}, initializes complex weights by independently sampling their real and imaginary parts using the real-valued Xavier bounds, under the assumption that they are uncorrelated and identically distributed. This method ensures that the total variance of each complex weight remains consistent with the target variance, though it does not enforce isotropy in the complex plane.

\subsubsection{Kaiming He Weight Initialization}

Xavier initialization scheme met with problems with the ReLU activation function, causing a single layer to have a higher standard deviation on average (of the order of the square root of the number of input connections divided by the square root of two). This led Kaiming He to propose an initialization scheme, which involves sampling the weights for a layer from a normal distribution \cite{he2015delving}. The standard deviation of the normal distribution depends on the number of input nodes ($n$). In particular, the weights are sampled from the distribution:

$$Z \sim N \left(0, \cfrac{2}{n} \right); \; f_Z (z) = 
\cfrac{\sqrt{n}}{2 \sqrt{\pi}} \exp \left( - \cfrac{n x^2}{4} \right)
$$
The above algorithm can be extended for the complex normal random variable by noting that the probability density function of a complex random vector Z is the joint density function of its real and imaginary parts. The proposed distribution is circular-symmetric owing to the independent real and imaginary components having zero mean and the same variance:
$$f_Z (z) = 
\cfrac{1}{(2 \pi) \sqrt{\det (\Sigma_Z)}} \exp \left( -\cfrac{1}{2} (z - \bm{0})^\top \Sigma_Z^{-1} (z - \bm{0}) \right)
$$
$$ =
\cfrac{1}{(2 \pi) \sqrt{\det 
\begin{pmatrix}
2/n & 0 \\
0 & 2/n \\
\end{pmatrix}
}} \exp \left( -\cfrac{1}{2}
\begin{pmatrix}
    x \\
    y \\
\end{pmatrix}
^\top
\begin{pmatrix}
2/n & 0 \\
0 & 2/n \\
\end{pmatrix}^{-1}
\begin{pmatrix}
    x \\
    y \\
\end{pmatrix}\right)
$$ $$ = 
\cfrac{n}{4 \pi} \exp \left( -\cfrac{n (x^2 + y^2)}{4} \right)
= 
\cfrac{n}{4 \pi} \exp \left( -\cfrac{n \left\| z \right\|^2}{4} \right)
$$
The sampling process for complex-valued parameters follows the same foundational principles as those used for bivariate real-valued distributions.

Building upon the general initialization schemes discussed earlier, a notable advancement was described by Trabelsi et al. \cite{trabelsi2018deep}, who presented a theoretically grounded initialization method designed to preserve the scale of activations and gradients across layers for complex-valued neural networks. In their method, the complex weight parameter $W$ is initialized by separately sampling its magnitude and phase. The magnitude of $W$ is drawn from a Rayleigh distribution with a mode parameter $\sigma$, which is carefully derived to control the variance of the resulting complex weight. The phase component is sampled independently from a uniform distribution over the interval $[-\pi, \pi]$. The final complex weight is then constructed by multiplying the sampled magnitude by the phasor $e^{i\theta}$, where $\theta$ is the sampled phase. This initialization ensures a balanced distribution of complex weights in the Argand plane while maintaining a well-defined variance, thereby promoting gradient flow and improving training stability in deeper architectures. This concludes our discussion on some of the theoretical underpinnings that form the bedrock of CVCNNs. The theoretical frameworks presented in this section collectively serve as the scaffolding upon which the subsequent empirical investigations and applications of CVCNNs will be built.

\section{Experiments}\label{ch:experiments}

We start by investigating the performance of CVCNNs on standard image datasets to gain insights into their behavior in comparison to real-valued networks. Following this, we shift our focus towards audio applications, specifically music genre classification. This transition allows us to explore the adaptability and effectiveness of CVCNNs in a domain where complex-valued representations hold potential advantages. By examining both visual and auditory data, we aim to comprehensively understand the capabilities and limitations of CVCNNs across different domains.

\subsection{Experiment 1: Comparing CVCNNs on Standard Image Datasets}

In this experiment, we evaluate the performance of CVCNNs on standard image datasets, namely MNIST, FMNIST, and KMNIST. MNIST \cite{lecun1998mnist} consists of handwritten digits (0–9) and serves as a classical benchmark for image classification. Fashion-MNIST \cite{xiao2017fashionmnist} is a drop-in replacement for MNIST, containing grayscale images of clothing items. Kuzushiji-MNIST \cite{clanuwat2018kmnist} introduces cursive Japanese characters, offering a more challenging variant with cultural and linguistic diversity. The objective is to understand how CVCNNs handle image data compared to traditional real-valued neural networks.

\textbf{Method:} We implement a baseline CNN architecture designed for image classification, comprising two convolutional layers followed by batch normalization, max pooling, and two fully connected layers. ReLU activations are applied after each convolutional layer, and the final output is computed using the absolute value followed by a log softmax function. This real-valued CNN serves as the reference model for comparison. To explore the impact of complex representations, we construct a corresponding CVCNN using the same overall structure, with all operations extended to the complex domain. In this version, the ReLU activations are replaced by the $\mathbb{C}\text{ReLU}$ function, which applies ReLU separately to the real and imaginary parts. Complex batch normalization is employed, and weight initialization follows the method described by Trabelsi et al. \cite{trabelsi2018deep}. To evaluate performance across varying input configurations, we consider five different experimental setups:
\begin{enumerate}
    \item Real-Valued CNN (Baseline): The input consists of standard real-valued image data with no imaginary component.
    \item CVCNN (Real Valued Input): A complex-valued CNN architecture is used, but the input contains only the real component (i.e., the imaginary part is zero).
    \item CVCNN (Fixed Imaginary Component): The real-valued image data (training set) is augmented with a constant imaginary component of 0.1 across all pixels.
    \item CVCNN (Fixed Phase Input): The real-valued image data (training set) is augmented with a constant phase component of 0.5 radians across all pixels, maintaining the magnitudes.
    \item CVCNN (Random Complex Perturbation): The real-valued image data (training set) is augmented with small random values (of order $10^{-5}$) added to both real and imaginary parts across all pixels.  
\end{enumerate}

These variations are designed to probe the behavior of CVCNNs under controlled conditions that selectively activate the complex structure of the network. Comparing settings 1 and 2 isolates the architectural impact of using a complex-valued network even when no imaginary information is provided. Setting 3 introduces controlled imaginary information to investigate how consistent imaginary offsets affect learning. Setting 4 explores the importance of phase as a distinct informational signal, and setting 5 examines the network’s sensitivity to noisy or unstable phase and magnitude fluctuations. By analyzing model performance under these diverse conditions, we aim to better understand the benefits and limitations of complex-valued representations in neural networks and how they interact with phase, magnitude, and structure in input data. We measure training and testing accuracy for every dataset, along with the time taken for each epoch in the training process. Since all the datasets used in our experiments are approximately class-balanced, accuracy serves as an appropriate and interpretable performance metric. Tracking both training and test accuracy across epochs provides a clear view of the learning curve, helping us assess not only peak performance but also convergence behavior, generalization, and potential overfitting or underfitting trends over time.

\begin{figure}[h]
  \centering
  \begin{subfigure}{\linewidth}
    \centering
    \includegraphics[width=\linewidth]{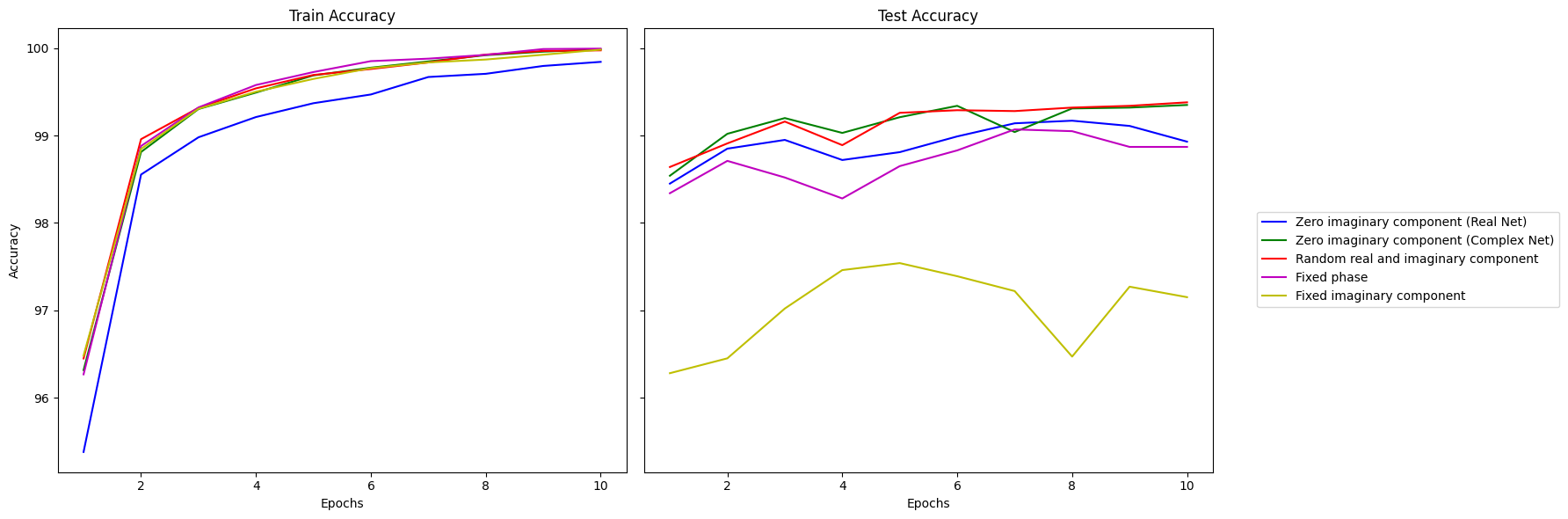}
    \caption{Experiment 1 Results on MNIST Dataset}
    \label{fig:mnist}
  \end{subfigure}

  \vspace{8pt} 

  \begin{subfigure}{\linewidth}
    \centering
    \includegraphics[width=\linewidth]{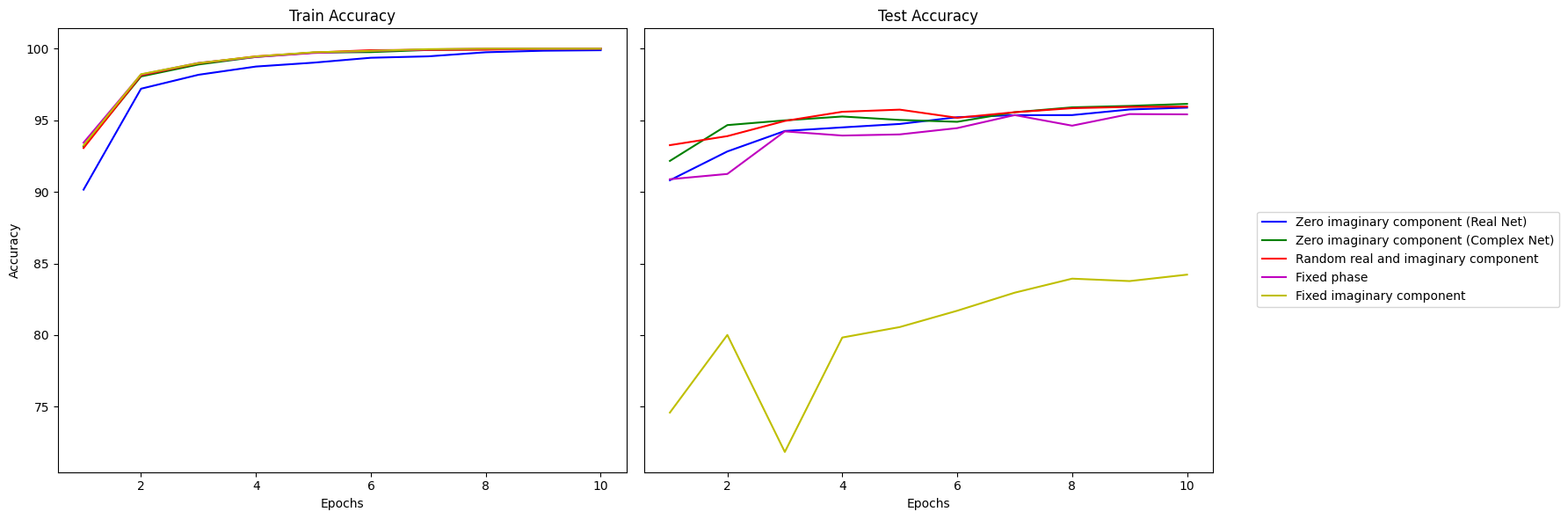}
    \caption{Experiment 1 Results on KMNIST Dataset}
    \label{fig:kmnist}
  \end{subfigure}

  \vspace{8pt}

  \begin{subfigure}{\linewidth}
    \centering
    \includegraphics[width=\linewidth]{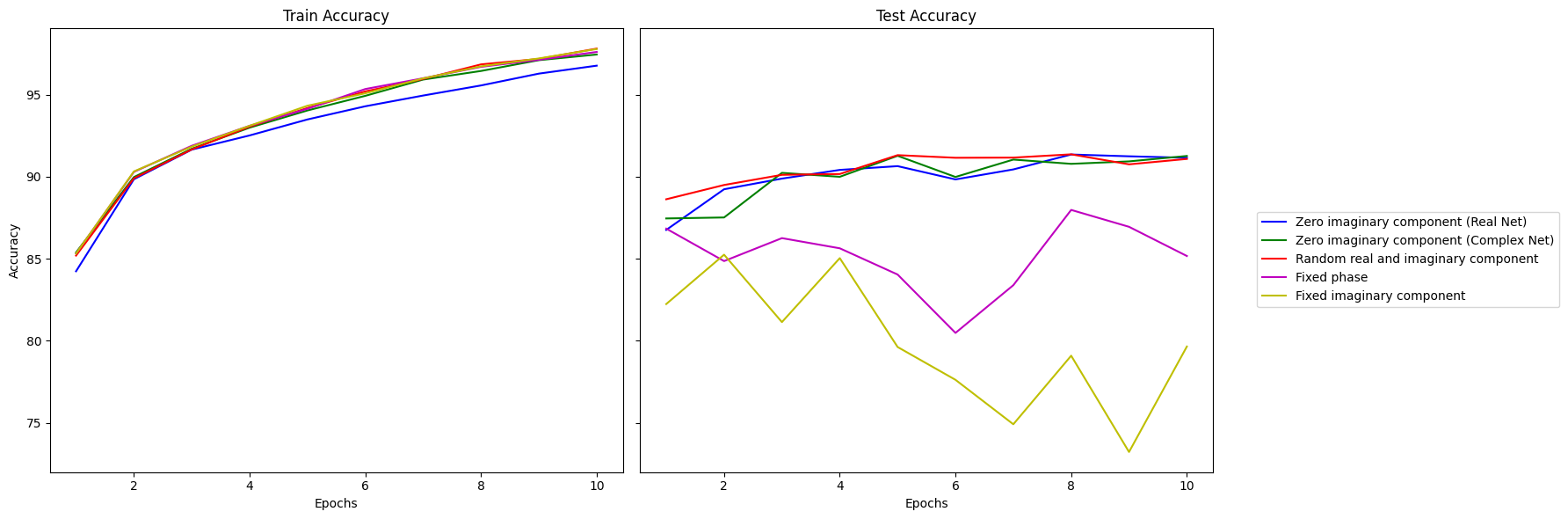}
    \caption{Experiment 1 Results on FMNIST Dataset}
    \label{fig:fmnist}
  \end{subfigure}

  \caption{Experiment 1 Results Across Three Datasets}
  \label{fig:e1_all_results}
\end{figure}

The experiment results, presented in Figure \ref{fig:e1_all_results}, demonstrate that CVCNNs achieve comparable performance to their real-valued counterparts across the MNIST, KMNIST, and FMNIST datasets. The training and testing accuracies remain nearly identical in most settings, suggesting that CVCNNs are capable of learning effectively even in the absence of a strong imaginary signal (Setting 2). When synthetic phase components are introduced (Setting 4), no severe degradation is observed, indicating that the model can handle phase manipulations without significant loss in performance. This is likely because phase affects the directional structure of features in a way that preserves overall signal coherence, allowing the network to adapt without disrupting key representational patterns.  However, in Setting 3 (Fixed Imaginary Component), slight performance drops in the test accuracy are noted, particularly for KMNIST and Fashion-MNIST, suggesting that injecting synthetic imaginary parts may interfere with the learning process on more complex visual patterns. 

This degradation likely stems from the fact that these transformations were applied only during training, while the test data remained in its original form. As a result, the model learns to rely on features present only in the transformed training data, creating a distribution shift that affects generalization, especially when the fixed imaginary parts introduce unnatural or non-informative patterns. In contrast, modifications to phase may not create such a sharp mismatch, as these tend to preserve overall structure or can be learned in a way that generalizes better to the unmodified test set. Nonetheless, in Setting 5 (Random Complex Perturbation), the model remains robust, implying that CVCNNs are not overly sensitive to small, unstructured perturbations in the complex plane and maintain good generalization. Finally, across all datasets, training times for CVCNNs were significantly higher, about 4-5 times more than their real counterparts, owing to the additional complexity of handling operations in the complex domain. It is important to note that these experiments primarily assess the architectural competence and stability of CVCNNs under controlled synthetic manipulations. While no significant performance advantage is observed under these toy configurations, the ability of CVCNNs to remain stable and competitive suggests potential for more structured and phase-rich domains.

To further deepen our understanding of performance variations in Complex-Valued Convolutional Neural Networks (CVCNNs), we systematically investigate the impact of different complex activation functions on learning outcomes across a consistent experimental setup. Specifically, we compare the performance of five complex activations: the standard $\mathbb{C}$ReLU (as before), modReLU, the smoothed version of zReLU, split tanh activations, and the complex cardioid activation. Each of these activations is evaluated under four data transformation settings applied to the input: (1) no transformation (baseline), (2) small random complex noise, (3) fixed phase perturbation, and (4) fixed imaginary component. We use the same network architecture, training configuration, and optimizer settings across all activation-function experiments to ensure a fair comparison. Each model is trained for 5 epochs on the same datasets, and we report training and test accuracies in the corresponding performance tables to isolate the effect of the complex activation functions and understand their robustness to input transformations:

\begin{table}[htbp]
\centering
\caption{CVCNN Train and Test Accuracy (\%) on MNIST for Different Activations}
\begin{tabular}{lcccccccc}
\toprule
\textbf{Activation} 
& \multicolumn{4}{c}{\textbf{Train Accuracy (\%)}} 
& \multicolumn{4}{c}{\textbf{Test Accuracy (\%)}} \\
\cmidrule(lr){2-5} \cmidrule(lr){6-9}
& None 
& \makecell{Small\\Noise} 
& \makecell{Fixed\\Phase} 
& \makecell{Fixed\\Imaginary} 
& None 
& \makecell{Small\\Noise} 
& \makecell{Fixed\\Phase} 
& \makecell{Fixed\\Imaginary} \\
\midrule
CReLU         & 98.90 & 98.94 & 98.94 & 98.95 & 99.07 & 99.01 & 94.06 & 98.28 \\
modReLU       & 96.73 & 96.51 & 96.68 & 96.60 & 97.01 & 97.49 & 71.84 & 58.97 \\
smooth zReLU  & 93.23 & 98.86 & 98.89 & 98.89 & 81.07 & 98.88 & 44.31 & 98.02 \\
split tanh    & 98.16 & 98.15 & 98.10 & 98.19 & 98.23 & 98.26 & 63.96 & 88.26 \\
cardioid      & 99.12 & 99.13 & 99.08 & 99.15 & 99.14 & 99.09 & 93.54 & 98.80 \\
\bottomrule
\end{tabular}
\label{t1}
\end{table}

\begin{table}[htbp]
\centering
\caption{CVCNN Train and Test Accuracy (\%) on KMNIST for Different Activations}
\begin{tabular}{lcccccccc}
\toprule
\textbf{Activation} 
& \multicolumn{4}{c}{\textbf{Train Accuracy (\%)}} 
& \multicolumn{4}{c}{\textbf{Test Accuracy (\%)}} \\
\cmidrule(lr){2-5} \cmidrule(lr){6-9}
& None 
& \makecell{Small\\Noise} 
& \makecell{Fixed\\Phase} 
& \makecell{Fixed\\Imaginary} 
& None 
& \makecell{Small\\Noise} 
& \makecell{Fixed\\Phase} 
& \makecell{Fixed\\Imaginary} \\
\midrule
CReLU         & 98.34 & 98.32 & 98.33 & 98.27 & 94.87 & 95.17 & 86.02 & 93.87 \\
modReLU       & 93.48 & 93.80 & 93.60 & 93.61 & 88.11 & 88.63 & 46.78 & 35.52 \\
smooth zReLU  & 98.25 & 98.21 & 98.20 & 98.27 & 95.05 & 94.95 & 55.48 & 92.62 \\
split tanh    & 96.17 & 96.18 & 96.06 & 96.19 & 91.04 & 90.69 & 57.44 & 70.89 \\
cardioid      & 98.60 & 98.57 & 98.65 & 98.61 & 95.29 & 95.31 & 81.80 & 94.21 \\
\bottomrule
\end{tabular}
\label{t2}
\end{table}

\begin{table}[htbp]
\centering
\caption{CVCNN Train and Test Accuracy (\%) on FMNIST for Different Activations}
\begin{tabular}{lcccccccc}
\toprule
\textbf{Activation} 
& \multicolumn{4}{c}{\textbf{Train Accuracy (\%)}} 
& \multicolumn{4}{c}{\textbf{Test Accuracy (\%)}} \\
\cmidrule(lr){2-5} \cmidrule(lr){6-9}
& None 
& \makecell{Small\\Noise} 
& \makecell{Fixed\\Phase} 
& \makecell{Fixed\\Imaginary} 
& None 
& \makecell{Small\\Noise} 
& \makecell{Fixed\\Phase} 
& \makecell{Fixed\\Imaginary} \\
\midrule
CReLU         & 90.92 & 90.79 & 90.90 & 90.78 & 89.76 & 89.73 & 78.37 & 86.15 \\
modReLU       & 86.32 & 86.13 & 86.21 & 86.41 & 86.45 & 86.41 & 74.14 & 82.99 \\
smooth zReLU  & 90.41 & 90.50 & 90.45 & 90.54 & 88.92 & 89.44 & 66.62 & 87.98 \\
split tanh    & 88.17 & 88.37 & 88.15 & 88.39 & 87.04 & 87.88 & 79.93 & 84.66 \\
cardioid      & 91.99 & 92.16 & 92.02 & 91.85 & 90.09 & 90.31 & 78.55 & 86.74 \\
\bottomrule
\end{tabular}
\label{t3}
\end{table}

As shown in Tables \ref{t1}, \ref{t2}, and \ref{t3}, the cardioid activation consistently yields top-tier performance across all three datasets, achieving the highest or near-highest test accuracy, particularly in the presence of complex-valued transformations. It remains robust under both phase and imaginary distortions, highlighting its adaptability to varying input characteristics. In contrast, modReLU shows notable instability under fixed phase and imaginary constraints, with steep drops in test accuracy (e.g., 46.78\% on KMNIST and 74.14\% on Fashion-MNIST), suggesting it may be sensitive to rigid complex structure manipulations. Interestingly, the smooth version of zReLU performs competitively under untransformed and noisy settings but deteriorates sharply in the fixed phase scenario for KMNIST and FMNIST. Split tanh, while generally more stable than modReLU and zReLU, also shows susceptibility under stronger complex constraints, though to a lesser degree. CReLU remains a strong baseline, offering solid and relatively consistent results, but is occasionally outperformed by cardioid in terms of generalization. These results suggest that activations like cardioid, which are natively phase-aware, offer better generalization in complex-valued settings, especially when data undergoes non-trivial complex transformations.

\subsection{Experiment 2: MFCCs for Music Genre Classification}

In this experiment, we shift our focus to the domain of audio signal processing, specifically targeting the task of music genre classification. We use the GTZAN dataset \cite{tzanetakis2002gtzan}, a widely adopted benchmark for this task, which consists of 1000 audio tracks categorized into 10 genres such as classical, jazz, pop, and rock.
Each track is 30 seconds long and sampled at 22,050 Hz, making the dataset suitable for evaluating models on genre recognition from raw or transformed audio signals. The choice of the music genre classification task in the domain of audio signal processing was made for several reasons.

Firstly, it serves as a foundational and relatively simple task within the broader field of audio signal processing. Music genre classification involves categorizing audio clips into predefined genres based on their acoustic characteristics. This task provides a clear and well-defined objective, making it suitable for experimentation and evaluation of different approaches. Furthermore, the relevance of music genre classification extends across various industries and applications. Content recommendation systems heavily rely on accurately classifying music into genres to provide users with personalized playlists and suggestions.  Traditionally, real-valued approaches have been employed, primarily relying on supervised feature extraction techniques and frequency domain representations, such as Mel-Frequency Cepstral Coefficients (MFCCs), to tackle this problem. In this work, we will draw on some of these established approaches while also exploring the transformative potential of CVCNNs in this domain.

\begin{figure}[H]
  \centering
\includegraphics[width = \linewidth]{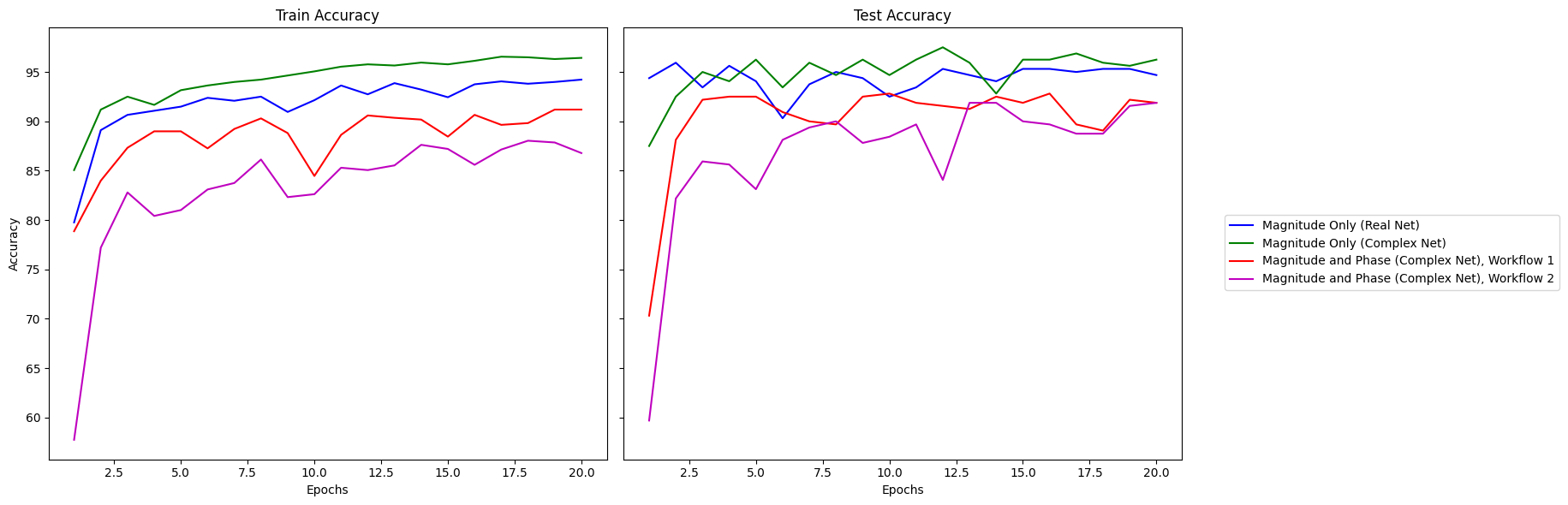}
  \caption{Experiment 2 Results (Binary Music Genre Classification)}
  \label{figexp2res}
\end{figure}

A key consideration in these methods is how the audio signal is represented, since representation fundamentally shapes the features available for classification. In audio signal processing, representation can take place in the time domain or the frequency domain.  Time-domain representation encapsulates the amplitude of a signal as a function of time, providing insights into the waveform's temporal characteristics. Conversely, frequency-domain representation decomposes the signal into its constituent frequencies, revealing the underlying spectral content. At the heart of frequency-domain representation lies the Fourier Transform, a mathematical operation that takes the signal as input and decomposes it into a sum of sine and cosine waves of varying frequencies, having their amplitude and phase. The resulting representation is what constitutes the frequency spectrum. For real-world signals like audio, we typically use the Short-Time Fourier Transform (STFT) to capture both time and frequency information. STFT applies the Fourier Transform on short overlapping segments of the signal using a window function $\phi(t)$.  Given a discrete-time signal $x[i]$, a window function $\phi(i - \tau)$, and window size $N$, the STFT at frequency bin $k$ and time index $\tau$ is defined as:
$$\hat{h}(k, \tau) = \sum_{i = 0}^{N-1} x[i] \phi(i - \tau) \exp\left(-2\pi i \cdot \frac{k}{N}\right)$$
This yields a complex output, where the magnitude represents the amplitude and the angle represents the phase of the frequency component. However, for most applications, we care only about the magnitude of the transformation and simply ignore the associated phase. In this context, CVCNNs, engineered to operate on complex-valued data, offer an elegant solution for harnessing the full potential of the complex spectrum obtained from the Fourier Transform. The spectral representation arising from STFT forms the basis for the extraction of MFCCs, which involves a multi-step procedure:
\begin{itemize}
    \item Frame-by-Frame Segmentation: The audio signal is partitioned into short overlapping frames, typically spanning around 20-40 milliseconds, essential for capturing localized spectral information.
    \item Windowing: Within each frame, a window function is applied to emphasize the central portion and mitigate spectral leakage effects. For our experiments, we have used the Hann window function, which does a cosine transform on the signal:
    $$\phi(i) = \cfrac{1}{2} \left(1 - \cos \left( \cfrac{2 \pi i}{n_f - 1} \right) \right)$$
    where $n_f$ is the number of samples in a frame. Other popular choices of window functions include Gaussian, Welch, and Spline functions.
    \item Fourier Transform: The Fourier Transform is then employed to convert the signal into the frequency domain, yielding the magnitude spectrum.
    \item Mel Filtering: The human auditory system does not perceive frequencies linearly. To align feature extraction with perceptual characteristics, a series of overlapping triangular filters in the Mel frequency scale is applied to the magnitude spectrum \cite{Karjalainen1988SpeechCH}. The mapping from frequency $f$ (in Hz) to the Mel scale is defined as
    $$m = 2595 \log_{10} \left( 1 + \cfrac{f}{700} \right)$$
    The filters serve to emphasize lower frequencies, which are more perceptually relevant, while attenuating higher frequencies. The mel filters are computed as follows:
    $$H_m(i) = \left\{
    \begin{array}{cc}
        0 &  i < f(m - 1)\\
        \cfrac{i - f(m - 1)}{f(m) - f(m - 1)} & f(m - 1) \leq i < f(m) \\
        1 & i = f(m) \\
        \cfrac{f(m + 1) - i}{f(m + 1) - f(m)} & f(m) < i \leq f(m + 1) \\
        0 & i > f(m + 1)
    \end{array} \right\}$$
    where $f(m)$ represents the center frequency (in terms of STFT bin index) of the $m^\text{th}$ Mel filter after converting Mel-scaled frequency back to Hz and mapping it to the appropriate discrete STFT bin.
    \item Logarithmic Compression: The logarithm of the filterbank energies is computed, reflecting the logarithmic perception of loudness in the human auditory system.
    \item Discrete Cosine Transform (DCT): Finally, the application of DCT decorrelates the Mel-filterbank coefficients, resulting in a compact representation that captures the spectral characteristics via the cepstral coefficients.
\end{itemize}
The resulting MFCCs serve as an informative yet concise representation of the audio signal, capturing both temporal dynamics and spectral content. Incorporating CVCNNs into the MFCC processing pipeline augments the feature extraction process. Rather than discarding the phase information after taking the STFT, which can contain valuable insights about the audio, CVCNNs retain the entirety of the complex spectrum. This requires some modifications to the MFCC extraction pipeline to preserve the phase values to avoid discarding them during the power spectrum calculation. We explore two ways of doing it (Figure \ref{figcompmfccwf}):
\begin{figure}[h]
  \centering
\includegraphics[scale = 0.4]{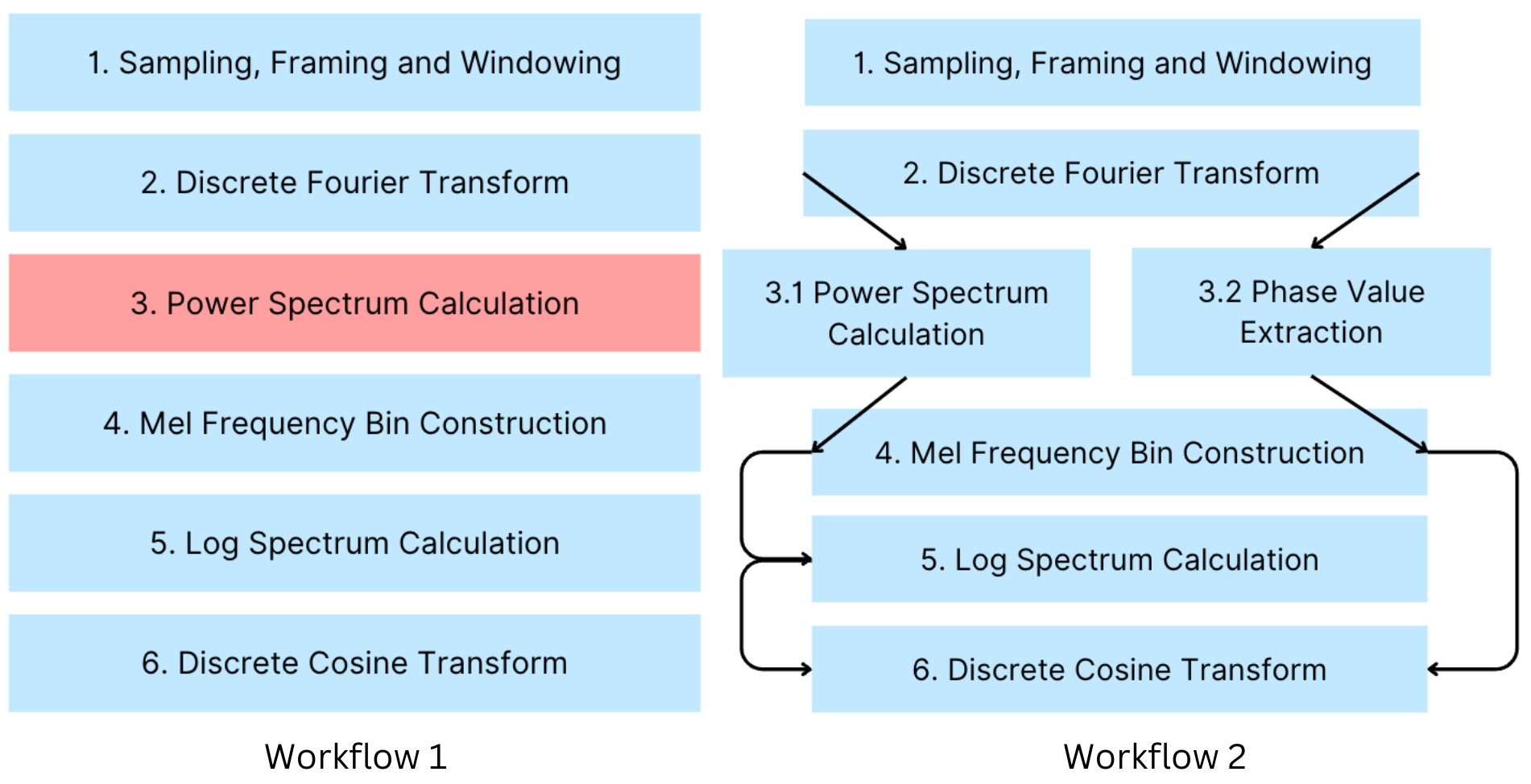}
  \caption{Experiment 2 Complex Valued MFCC Extraction Workflows}
  \label{figcompmfccwf}
\end{figure}
\begin{enumerate}
    \item Skip power spectrum calculation: After computing the STFT, we bypass the conversion to a power spectrum and retain the raw complex-valued STFT output. This allows the CVCNN to learn directly from both magnitude and phase components without applying any lossy transformation.
    \item Extract and reintegrate phase: After computing the STFT, we separately extract the phase information and retain it. The standard MFCC pipeline proceeds on the magnitude spectrum, and the saved phase is later integrated after the log spectrum stage. This ensures that traditional MFCC features are preserved while still enabling the network to access the complementary phase information.
\end{enumerate}
\textbf{Method:} We implement a simple baseline CNN architecture tailored for binary audio genre classification tasks (Rock vs Classic), consisting of two convolutional layers, each followed by activation,  max pooling, batch normalization, and two fully connected layers along with activation. Instead of using all 10 genres from the original dataset, we focus on a binary classification task (Rock vs Classic) to simplify the experimental setup and isolate the effects of architectural and input variations, making it easier to interpret performance differences across model configurations. As before, ReLU activation functions are used after each convolutional layer, and the final output is obtained by using the absolute value followed by a log softmax function. We measure both training and testing accuracy for each of the following methods:
\begin{enumerate}
    \item Real-valued MFCCs with a real CNN
    \item Real Valued MFCCs with a CVCNN
    \item Complex Valued MFCCs (Workflow 1) with a CVCNN
    \item Complex Valued MFCCs (Workflow 2) with a CVCNN
\end{enumerate}
The experiment results are presented in Figure \ref{figexp2res}. For the training accuracy, it is evident that the utilization of real-valued MFCCs in conjunction with CVCNNs yields the most promising outcomes, superseding even the application of the same data with real-valued CNNs. Interestingly, the performance of complex-valued MFCCs on the training set registers a relatively diminished performance across both workflows. However, the assessment of test set performance indicates a convergence of results across all methodologies as the number of epochs increases, suggesting a resilience of CVCNNs towards overfitting when provided with complex-valued MFCCs. Analogous trends are observed in the trajectories of both training and test losses.
Just as in Experiment 1, we also compare the performance of the complex-valued network on the original input data and across the two workflows for different choices of complex activation functions.

\begin{table}[htbp]
\centering
\caption{CVCNN Train and Test Accuracy (\%) on Binary Music Genre Classification for Different Activations}
\begin{tabular}{lcccccc}
\toprule
\textbf{Activation} 
& \multicolumn{3}{c}{\textbf{Train Accuracy (\%)}} 
& \multicolumn{3}{c}{\textbf{Test Accuracy (\%)}} \\
\cmidrule(lr){2-4} \cmidrule(lr){5-7}
& \makecell{Magnitude \\ Only}  
& \makecell{Magnitude \\ \& Phase \\ (Workflow 1)} 
& \makecell{Magnitude \\ \& Phase \\ (Workflow 2)}   
& \makecell{Magnitude \\ Only}  
& \makecell{Magnitude \\ \& Phase \\ (Workflow 1)} 
& \makecell{Magnitude \\ \& Phase \\ (Workflow 2)}  \\
\midrule
CReLU         & 92.24 & 85.39 & 76.26 & 92.31 & 90.28 & 78.78 \\
modReLU       & 52.77 & 80.88 & 52.74 & 50.66 & 82.56 & 54.34 \\
zReLU         & 90.90 & 86.68 & 76.88 & 92.00 & 88.09 & 78.06 \\
split tanh    & 91.21 & 77.06 & 55.20 & 90.84 & 79.66 & 52.72 \\
cardioid      & 93.18 & 86.90 & 79.71 & 95.34 & 89.75 & 84.75 \\
\bottomrule
\end{tabular}
\label{t4}
\end{table}

Table \ref{t4} summarizes the performance of the CVCNN model trained over 10 epochs using five different complex activation functions across the three data configurations (magnitude only and the two workflows). For comparison, the real-valued CNN with the same architecture achieved a test accuracy of 92.5\% and a train accuracy of 92.14\%, serving as a strong baseline.

A few consistent patterns emerge across the datasets. Firstly, CVCNNs trained on real-valued MFCCs (i.e., magnitude only) perform the best overall, especially when combined with smoother or bounded complex activations like the cardioid or zReLU. This may be attributed to the architectural limitations in handling phase information effectively; the network is not explicitly designed to disentangle or exploit phase relationships, which could require specialized modules or phase-aware convolutions. However, this does not diminish the potential usefulness of phase; instead, it highlights the need for more expressive designs.

Importantly, certain activations such as modReLU underperform severely in some configurations, particularly in Workflow 2, where the accuracy drops close to random ($\sim$50\%). This is primarily due to vanishing or unstable gradients caused by the activation’s steep non-linearity and its sensitivity to input magnitude, which can make optimization highly challenging. Prior work has similarly observed that the unboundedness of certain activation functions in complex-valued neural networks can lead to numerical instabilities, including practically infinite gradients and training collapse when the optimization process encounters singularities or steep regions of the activation landscape \cite{monning2018evaluation}. In such cases, recovering the training process becomes extremely difficult unless additional constraints such as weight normalization or gradient clipping are applied. Motivated by these observations, we adopt aggressive gradient clipping in select configurations to stabilize learning and mitigate such pathological behaviors. 

It is also seen that activations like cardioid not only offer smoother training but also deliver strong performance across all workflows, even with complex inputs, achieving up to 95.34\% test accuracy. These findings align with our earlier experiments on benchmark image datasets, where the cardioid activation consistently outperformed other alternatives across most settings. A similar trend was reported by \cite{virtue2017better} in the context of MRI signal modeling, where their proposed complex cardioid function outperformed split sigmoid and other nonlinear activations by preserving phase information, leading to more stable and accurate identification of tissue parameters. These results underscore that while current CVCNN architectures are already competitive with real-valued baselines, especially for magnitude-only data, the design of suitable activation functions and phase-aware modules remains key to unlocking the full potential of complex representations.

\subsection{Experiment 3: Graph Neural Networks for Incorporating Phase Information}

In this section, we temporarily shift focus away from the architecture-specific analysis of CVCNNs to investigate a more fundamental question: To what extent is phase information intrinsically valuable in audio signal representations? While previous experiments evaluated the performance of CVCNNs with different types of complex-valued inputs, including those augmented with fixed or random phase components, we observed that the inclusion of phase information did not consistently result in significant improvements in classification accuracy for audio tasks. This raises the question of whether phase information itself carries meaningful discriminative content or whether its contribution is marginal in certain contexts. This inquiry is pursued through Graph Neural Networks (GNNs) to help examine inter-dependencies within the MFCC representation. Notably, Mönning \& Manandhar (2018) also caution that complex-valued networks tend to benefit most when the data is inherently complex or meaningfully projected to the complex domain, highlighting potential inefficiencies when this structure is absent \cite{monning2018evaluation}.

GNNs are a powerful tool in deep learning, tailored for tasks involving graph-structured data and non-Euclidean domains. At their core, GNNs leverage message-passing algorithms to aggregate and exchange information between neighboring nodes in a graph, allowing for the propagation of information through the entire graph structure.
\begin{figure}[H]
  \centering
\includegraphics[scale = 0.27]{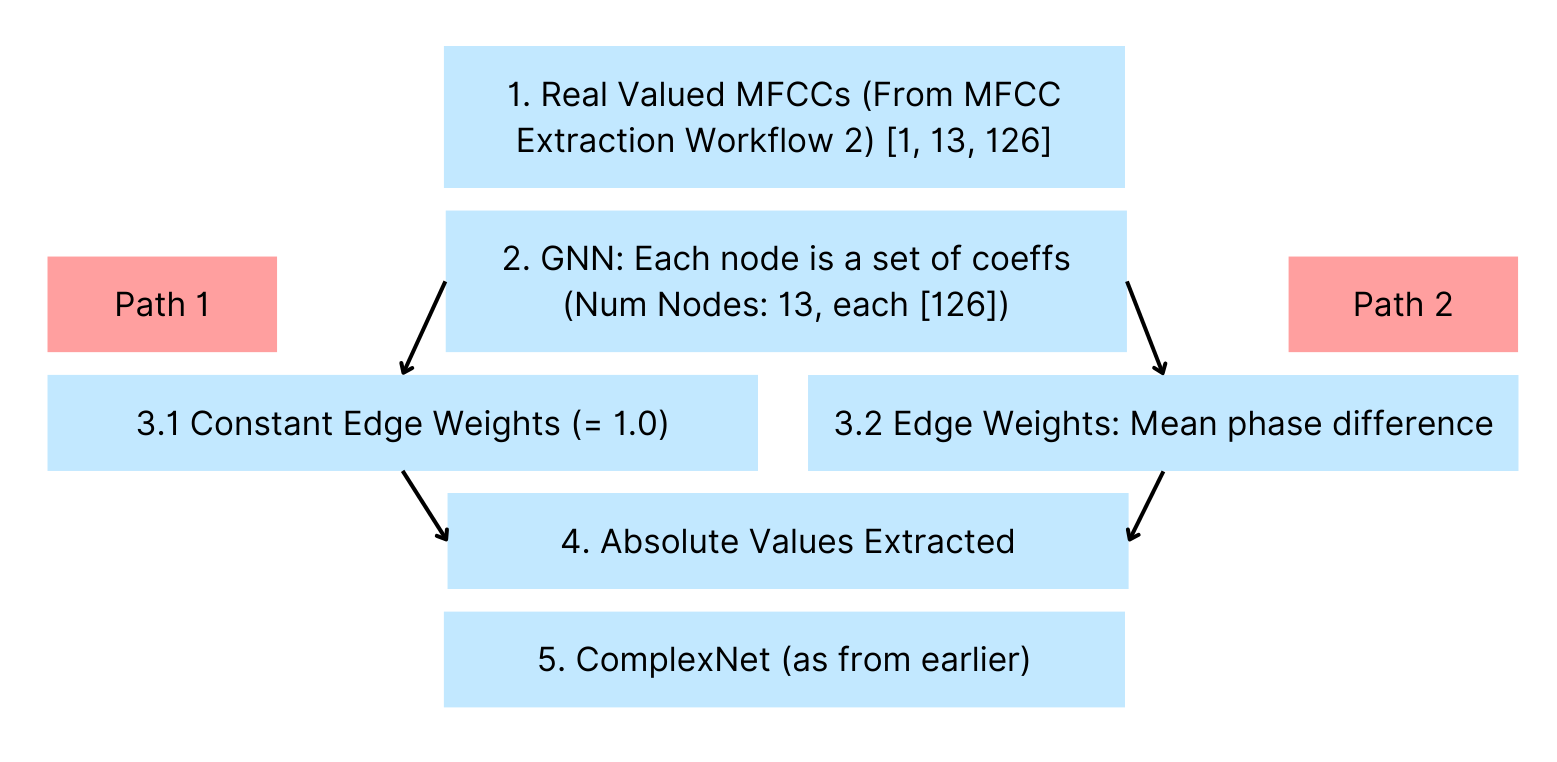}
  \caption{Experiment 3 GNN Workflows}
  \label{figexp3wf}
\end{figure}
\textbf{Method:} For the GNN, each vertex within the graph is represented by a vector comprising the $13$ MFCC coefficients, each vector being of dimensionality $126 \times 1$. These coefficients encapsulate critical spectral information learned from the audio signal. To evaluate the utility of phase information in audio analysis, we frame our experiment as a performance comparison between two distinct GNN-based workflows. Both workflows operate on graphs constructed from MFCC coefficient vectors extracted from the audio signal, which capture key spectral features. The core idea is to investigate whether incorporating phase information into the graph structure provides any meaningful advantage in classification performance. The two workflows differ in how they handle phase information within the graph (Figure \ref{figexp3wf}):
\begin{itemize}
    \item Workflow 1: Unweighted Graph (Excluding Phase Information): In this setup, phase information is entirely omitted. Each node in the graph represents an MFCC coefficient vector, and edges between nodes are uniformly weighted (or unweighted). This serves as a baseline where the GNN learns solely from the magnitude-based spectral features without any influence from phase.
    \item Workflow 2: Weighted Edges (Incorporating Phase Information): In contrast, this workflow incorporates phase information into the graph structure. Edges are weighted based on the mean absolute phase difference between the MFCC coefficient vectors of connected nodes. This introduces phase-aware connectivity, allowing the GNN to modulate information flow based on phase dissimilarity. 
\end{itemize}

The motivation for this comparative setup stems from the need to empirically isolate the contribution of phase information to the overall learning process. While CVCNNs implicitly encode phase through complex-valued computations, this GNN-based framework enables explicit modeling of phase as a structural feature within the graph. By contrasting the performance of the GNN in phase-agnostic and phase-aware settings, we can determine whether phase differences between temporal segments of audio data provide meaningful cues for classification.
\begin{figure}
  \centering
\includegraphics[width = \linewidth]{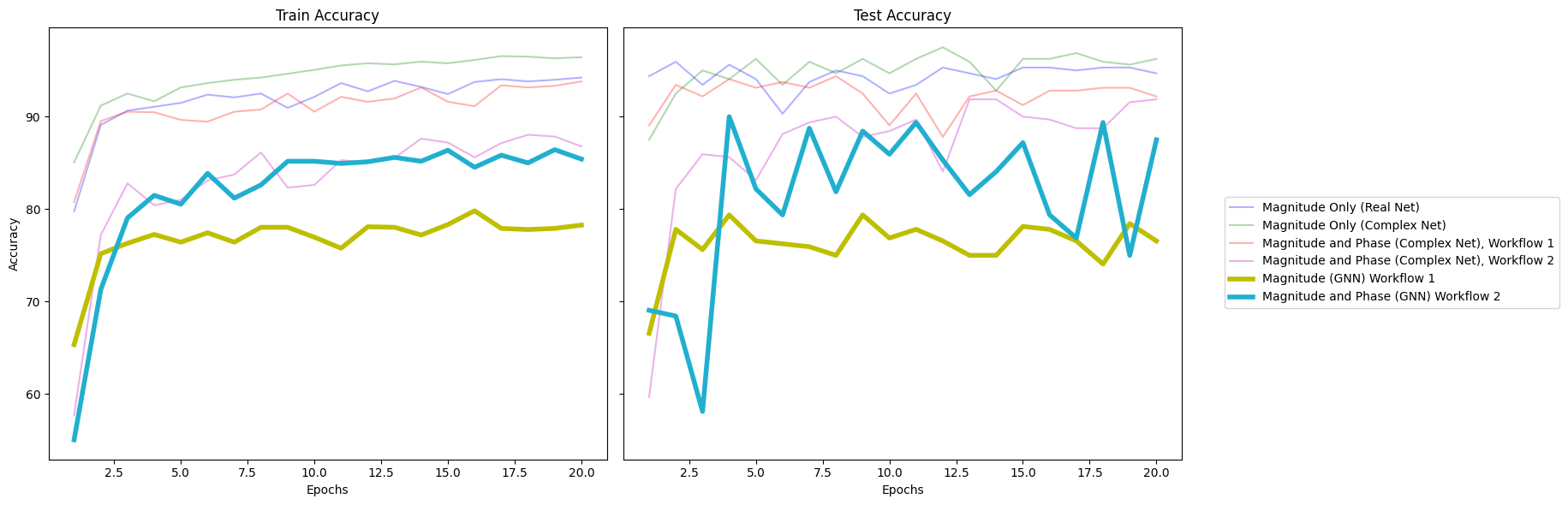}
  \caption{Experiment 3 Results (Binary Music Genre Classification)}
  \label{figexp3res}
\end{figure}
As depicted in Figure \ref{figexp3res}, a disparity in performance is evident between the two implemented GNN pipelines. Notably, the second workflow, which incorporates phase information, distinctly outperforms the first (for both train and test accuracy), wherein phase information is omitted. While it is noteworthy that both GNN architectures exhibit a performance level inferior to that of both the conventional neural networks and CVCNNs, their discrepancy underscores the substantive impact of phase values in the context of music genre classification.
\begin{figure}
  \centering
\includegraphics[width = \linewidth]{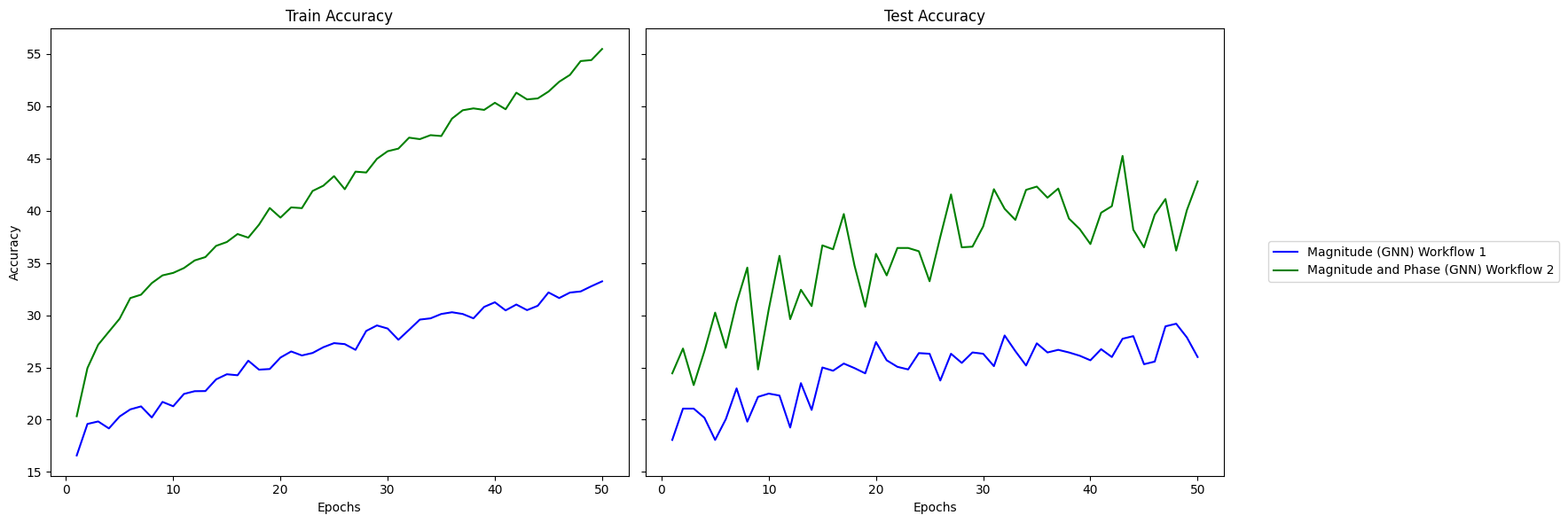}
  \caption{Experiment 3 Results (10-Class Genre Classification)}
  \label{figexp310c}
\end{figure}
Furthermore, as shown in Figure \ref{figexp310c}, it is observed that the trend extends consistently to the 10-class genre classification problem. The second workflow demonstrates markedly superior performance compared to the first, reaffirming the role of phase information in enhancing the discriminative capabilities of the classification model. While it is true that the current ways of harnessing phase information and integrating it within the framework of MFCC extraction don't yield superior performance relative to current research, it still sheds light on the importance of capturing the phase information and actively researching to find new pipelines and neural architectures that can utilize this information to make more informed decisions. And CVCNNs have a big part to play in this.

\section{Conclusion}

The exploration of CVCNNs in this study reveals their promise as a powerful extension to traditional deep learning architectures, especially in tasks that naturally involve phase-rich signals like audio. By developing a principled theoretical foundation, encompassing complex convolutions, activation functions, and normalization techniques, and testing their efficacy across both vision and audio domains, this work demonstrates that CVCNNs are not only mathematically elegant but also practically viable.

Experiments on standard image datasets confirm that CVCNNs can match real-valued models in performance, while remaining robust under complex-valued perturbations. More notably, in music genre classification tasks, these networks achieve test accuracies exceeding those of real-valued baselines when trained on real-valued MFCCs and show potential for improved generalization when phase information is appropriately integrated. The comparative analysis of complex activation functions further underscores the importance of thoughtful architectural choices; functions like the cardioid consistently deliver superior results across all experiments. However, the findings also reveal current architectural limitations in fully harnessing phase information, particularly in conventional CVCNN pipelines. While the phase-aware GNN-based representations show that phase can encode valuable discriminative signals, realizing its full potential will require more sophisticated strategies for encoding and utilizing it within learning frameworks.

Future research may focus on designing hybrid architectures that combine the expressive benefits of CVCNNs with structured phase-aware modules or self-attention mechanisms tailored for complex domains. The development of regularization strategies and training techniques to mitigate gradient instability, especially under non-analytic activations, remains an open area with high practical relevance. Additionally, deeper exploration into domains like biomedical imaging, communications, and acoustics may uncover new applications where the joint modeling of magnitude and phase can unlock richer representations. As deep learning continues to expand its reach, embracing the complex domain not only enhances the expressiveness of models but also challenges conventional assumptions about how we represent, learn from, and reason about data.

\bibliographystyle{plain}
\bibliography{references}  






\appendix

\section{Execution Environment and Setup}\label{apndA}

All experiments were conducted locally on a MacBook Pro (Model Number: MPHE3HN/A) running macOS 14.5 (Sonoma), equipped with an Apple M2 Pro chip featuring 10 CPU cores (6 performance + 4 efficiency) and 16 GB of unified memory. Training and evaluation were carried out using Python 3.11, with PyTorch and the complexPyTorch library for implementing complex-valued layers, activations, and batch normalization. Supporting packages included NumPy, SciPy, scikit-learn, Librosa, Matplotlib, and Seaborn for data processing, feature extraction (e.g., MFCCs), and visualization. All experiments were executed on the CPU, leveraging the Apple Silicon architecture for accelerated linear algebra operations via the PyTorch Metal backend.

\section{Access to Code Repository} \label{apndB}

For comprehensive access to the codes, models, and experimental setups utilized in this research, please refer to the dedicated repository hosted at \url{https://github.com/namanlab/UROPS_CVCNN}. Some components of the CVCNN architecture, such as complex convolution layers, complex batch normalization, weight initialization schemes, selected activation functions, and other complex-valued operations, were adapted from the open-source repository \url{https://github.com/wavefrontshaping/complexPyTorch/tree/master} \cite{complexPyTorch}.

\end{document}